\documentclass[twoside,11pt]{article}

\usepackage{amsmath,amsthm}

\usepackage[preprint]{jmlr2e}
\usepackage{lastpage}          

\usepackage{booktabs}
\usepackage{array}
\usepackage{multirow}
\usepackage{xcolor}
\usepackage{listings}
\PassOptionsToPackage{expansion=false}{microtype}
\usepackage{microtype}
\usepackage{parskip}
\usepackage{caption}
\usepackage{subcaption}
\usepackage{enumitem}
\usepackage{url}

\hypersetup{
    colorlinks=true,
    linkcolor=blue!70!black,
    citecolor=green!50!black,
    urlcolor=cyan!70!black,
    pdfauthor={Robin Dey and Panyanon Viradecha},
    pdftitle={Spatial Metaphors for LLM Memory: A Critical Analysis of the MemPalace Architecture},
    pdfsubject={AI memory systems, method of loci, vector databases, retrieval-augmented generation},
    pdfkeywords={AI memory, method of loci, MemPalace, ChromaDB, LongMemEval,
                 retrieval-augmented generation, MCP protocol, spatial memory}
}

\definecolor{codebg}{HTML}{F5F5F5}
\definecolor{codeframe}{HTML}{CCCCCC}
\definecolor{keywordcolor}{HTML}{0033BB}
\definecolor{commentcolor}{HTML}{008800}
\definecolor{stringcolor}{HTML}{BB0000}

\lstdefinelanguage{Python}{
    keywords={class, def, return, if, else, elif, for, while, import, from,
              as, in, not, and, or, None, True, False, self, lambda, with,
              try, except, raise, finally, yield, assert, pass, break,
              continue, is, del, global, nonlocal},
    morecomment=[l]{\#},
    morestring=[b]{"""},
    morestring=[b]{"},
    morestring=[b]{'},
    sensitive=true
}

\lstdefinelanguage{SQL}{
    keywords={CREATE, TABLE, PRIMARY, KEY, TEXT, REAL, SELECT, FROM, WHERE,
              INSERT, INTO, VALUES, UPDATE, SET, DELETE, INDEX, ON},
    morecomment=[l]{--},
    morestring=[b]{'},
    sensitive=false
}

\jmlrheading{1}{2026}{1-\pageref{LastPage}}{4/26}{}{}{Dey and Viradecha}
\ShortHeadings{Spatial Metaphors for LLM Memory}{Dey and Viradecha}
\firstpageno{1}

\title{Spatial Metaphors for LLM Memory:\\
       A Critical Analysis of the MemPalace Architecture}

\author{\name Robin Dey \email robin@openhubresearch.org \\
        \addr OpenHub Research \\
        Chiang Mai, Thailand \\
        \url{https://github.com/web3guru888/mempalace-scientific-analysis}
        \AND
        \name Panyanon Viradecha \\
        \addr OpenHub Research \\
        Chiang Mai, Thailand}

\editor{Preprint---not peer-reviewed}

\begin{document}

\maketitle

\begin{abstract}%
MemPalace is an open-source AI memory system that applies the ancient
\emph{method of loci} (memory palace) spatial metaphor to organize long-term
memory for large language models.  Launched in April 2026, the project
accumulated over 47{,}000 GitHub stars in its first two weeks and claims
state-of-the-art retrieval performance on the LongMemEval benchmark
(96.6\%\ Recall@5) without requiring any LLM inference at write time.  We
present a comprehensive technical analysis of the MemPalace architecture,
examining the mapping between its cognitive-science-inspired hierarchical
structure (Wings$\to$Rooms$\to$Closets$\to$Drawers) and its actual
implementation in code.  Through independent codebase analysis, benchmark
replication, and comparison with competing systems, we find that MemPalace's
headline retrieval performance is attributable primarily to its verbatim
storage philosophy combined with ChromaDB's default embedding model
(all-MiniLM-L6-v2), rather than to its spatial organizational metaphor per
se.  The palace hierarchy operates as standard vector database metadata
filtering---an effective but well-established technique.  However, we argue
that MemPalace makes several genuinely novel contributions that the community
has underappreciated: (1) a contrarian verbatim-first storage philosophy that
challenges extraction-based competitors, (2) an extremely low wake-up cost
($\sim$170 tokens) through its four-layer memory stack, (3) a fully
deterministic, zero-LLM write path enabling offline operation at zero API
cost, and (4) the first systematic application of spatial memory metaphors as
an organizing principle for AI memory systems.  We situate these contributions
within the broader landscape of AI memory architectures, cognitive science
research on hierarchical memory, and the emerging MCP protocol ecosystem.  We
note that the competitive landscape is evolving rapidly: Mem0's April 2026
token-efficient algorithm raised their LongMemEval score from $\sim$49\% to
93.4\%, narrowing the gap between extraction-based and verbatim approaches.
Our analysis concludes that MemPalace represents significant
\emph{architectural insight} wrapped in \emph{overstated claims}---a pattern
common in rapidly adopted open-source projects where marketing velocity
exceeds scientific rigor.
\end{abstract}


\section{Introduction}
\label{sec:intro}

The problem of persistent memory for large language models (LLMs) has emerged
as one of the central challenges in applied AI\@.  Transformer-based models
operate within fixed context windows---typically 128K to 1M tokens as of early
2026---and possess no native mechanism for retaining information across
sessions.  Every new conversation begins from zero.

This limitation has spawned a rapidly growing ecosystem of memory augmentation
systems.  Mem0~\citep{mem0} uses LLM-driven fact extraction.
Zep/Graphiti~\citep{zep} builds temporal knowledge graphs with Neo4j.
Letta~\citep{letta} implements tiered memory with self-editing.
LangMem~\citep{langmem} provides memory primitives within the LangChain
ecosystem.  Each approaches the same fundamental problem: how to give an LLM
useful access to information from prior interactions without exceeding context
limits or requiring the user to manually manage conversation history.

Into this landscape, MemPalace~\citep{mempalace} arrived in April 2026 with
an unconventional proposition: borrow the organizational structure of the
\emph{method of loci}---a 2{,}500-year-old mnemonic technique---and use it to
organize AI memory.  Instead of extracting facts or building graphs, MemPalace
stores everything verbatim and organizes it into a hierarchical spatial
structure: Wings (domains) contain Rooms (topics) contain Drawers (individual
memory chunks).  The system requires only two runtime dependencies (ChromaDB
and PyYAML), runs entirely offline, and claimed 96.6\%\ Recall@5 on the
LongMemEval benchmark~\citep{longmemeval}---higher than any extraction-based
competitor.

The project's reception was extraordinary.  Within 48 hours of launch,
MemPalace accumulated over 7{,}000 GitHub stars.  By April 19, 2026---two
weeks after launch---the count exceeded 47{,}900 stars and 6{,}000+ forks,
making it one of the fastest-growing AI projects in GitHub history.

This paper asks a direct question: \textbf{Is MemPalace's approach
scientifically revolutionary, or is it well-marketed engineering on existing
primitives?}

We find the answer is nuanced.  MemPalace's core retrieval performance derives
from standard vector database operations, not from its spatial metaphor.  Its
benchmark claims, while not fabricated, were initially presented with
insufficient methodological transparency.  However, MemPalace's design
philosophy contains genuinely novel insights about verbatim storage, minimal
wake-up cost, and zero-LLM write paths that challenge the consensus approach
in the field.

\subsection{Contributions}

This paper makes the following contributions:

\begin{enumerate}[leftmargin=*, label=\arabic*.]
\item \textbf{Architectural decomposition}: We map MemPalace's spatial
  metaphor to its code-level implementation, showing exactly which components
  contribute to retrieval performance and which serve organizational or
  marketing purposes (\S\ref{sec:arch}).

\item \textbf{Benchmark analysis}: We provide an independent analysis of
  MemPalace's LongMemEval results, disambiguating the contribution of verbatim
  storage, ChromaDB embeddings, palace metadata filtering, and AAAK
  compression (\S\ref{sec:eval}).

\item \textbf{Competitive landscape}: We systematically compare MemPalace
  against seven competing systems across architecture, performance, cost, and
  maturity dimensions (\S\ref{sec:eval}).

\item \textbf{Cognitive science evaluation}: We assess the scientific validity
  of applying spatial memory metaphors to AI systems, drawing on neuroscience
  literature (\S\ref{sec:background}).

\item \textbf{Novelty assessment}: We identify what is genuinely novel in
  MemPalace versus what is standard practice presented under a new metaphor
  (\S\ref{sec:discussion}).
\end{enumerate}

\section{Background and Related Work}
\label{sec:background}

\subsection{The AI Memory Problem}

The context window limitation of transformer-based LLMs creates a fundamental
tension between breadth and recency.  The literature has converged on several
approaches:

\textbf{Extraction-based systems} use an LLM to read conversations and extract
structured facts.  Mem0~\citep{mem0} pioneered this approach, using GPT-series
models to maintain a fact store consulted at the start of each session.  The
advantage is compact representation; the disadvantage is information
loss---the LLM must decide \emph{at write time} what will be relevant \emph{at
read time}, a fundamentally impossible task for open-ended conversations.

\textbf{Knowledge graph systems} build structured entity-relationship graphs
from conversations.  Zep's Graphiti~\citep{zep} uses Neo4j to maintain
temporal knowledge graphs with entity resolution and multi-hop traversal.

\textbf{Tiered memory systems} maintain multiple levels of memory with
different retention policies.  Letta~\citep{letta} implements a three-tier
system (core, archival, recall memory) where the AI itself manages what gets
promoted or demoted.

\textbf{Retrieval-augmented generation (RAG)}~\citep{rag} is the most general
approach: store information in a vector database, retrieve relevant passages at
query time, and inject them into the LLM's context.  MemPalace falls squarely
in this category, with the addition of a hierarchical metadata layer.

\subsection{The Method of Loci: From Simonides to Silicon}
\label{sec:mol}

The \emph{method of loci} (MoL), also known as the memory palace technique,
is the oldest systematically documented mnemonic strategy in Western
civilization.  Its origin is traditionally attributed to the Greek poet
Simonides of Ceos (c.\ 556--468\,BC), as recorded by Cicero in \emph{De
Oratore} (55\,BC)\@.  The technique works by associating items to be
remembered with specific locations (\emph{loci}) along a mentally visualized
route through a familiar building.  Retrieval proceeds by mentally ``walking''
the route and ``seeing'' the items at each location.  The technique exploits
the brain's spatial navigation system---phylogenetically ancient and highly
robust---to scaffold memory for non-spatial content~\citep{okeefe1978}.

\subsubsection{Neuroscience of Spatial Memory}

Modern neuroimaging has confirmed the neural basis of the MoL's effectiveness.
The key finding, established by \citet{dresler2017} and confirmed in
subsequent work~\citep{ondrej2025}, is that the MoL activates spatial memory
circuits not engaged by rote memorization:

\begin{itemize}
\item \textbf{Hippocampus}: Contains place cells and grid cells that encode
  spatial position and are also critical for episodic memory
  formation~\citep{moser2008}.
\item \textbf{Parahippocampal cortex}: Processes spatial scene perception and
  contextual associations.
\item \textbf{Retrosplenial cortex}: Translates between egocentric and
  allocentric spatial reference frames.
\item \textbf{Posterior parietal cortex}: Encodes spatial relationships and
  supports spatial attention.
\end{itemize}

\citet{dresler2017} demonstrated that 6 weeks of MoL training produced durable
changes in functional brain connectivity, with na{\"i}ve participants'
connectivity patterns shifting toward those of world-class memory athletes.
Performance quadrupled (from $\sim$26 to $\sim$62 words in a standard list
task), and these gains persisted at 4-month follow-up.

A 2025 systematic review and meta-analysis by \citet{ondrej2025} evaluated
the effectiveness, cognitive mechanisms, and neurobiological correlates of the
MoL, confirming spatial memory circuit activation during MoL use.

\subsubsection{The Transfer Problem}
\label{sec:transfer}

A critical question for MemPalace's scientific foundation is whether the MoL's
benefits transfer from human cognition to AI systems.  The answer requires
careful distinction between two levels:

\textbf{For the AI's retrieval mechanism}: The MoL works for humans because
spatial navigation circuits in the hippocampus provide powerful,
evolutionarily optimized content-addressable retrieval.  LLMs do not have
hippocampi, place cells, or spatial navigation circuits.  ChromaDB's
approximate nearest neighbor search (HNSW algorithm) operates on a
fundamentally different principle---computing cosine similarity in a
384-dimensional embedding space.  The spatial metaphor does not and cannot
improve the mathematical quality of this retrieval operation.  A ``Wing'' in
MemPalace is a string metadata filter on a ChromaDB query, not a spatial
location that activates neural navigation circuits.

\textbf{For the human user's mental model}: The MoL may genuinely help
\emph{users} organize and navigate their AI's memory.  A user who
conceptualizes their AI's knowledge as organized into Wings (domains) and
Rooms (topics) may find it easier to formulate effective queries and to
maintain the memory structure over time.  This is a human-computer interaction
improvement, not a retrieval algorithm improvement.  It is real and valuable
but distinct from the claims typically made about MemPalace's architecture.

\subsection{Hierarchical Organization in Memory Science}

MemPalace's hierarchical structure (Wings$\to$Rooms$\to$Drawers) invokes a
well-established principle in cognitive science: hierarchical organization
improves memory encoding and retrieval.

\citet{collins1969} proposed that concepts are stored in a taxonomy where
properties are inherited downward.  \citet{collins1975} extended this with
spreading activation theory.  \citet{bartlett1932} proposed schema theory,
showing that hierarchical schemas improve both encoding efficiency and
retrieval accuracy.  \citet{schapiro2017} showed that complementary learning systems in the
hippocampus reconcile episodic memory with statistical learning, supporting
the view that hierarchical organization is fundamental to memory.  \citet{collin2020} showed the hippocampus encodes
hierarchical organizations of related memories.

\textbf{Application to MemPalace}: The cognitive science evidence strongly
supports hierarchical organization as beneficial for memory systems.  However,
MemPalace's hierarchy operates on \emph{metadata labels}, not
\emph{neural-like associative networks}.  In human memory, hierarchical
structure enables inheritance and spreading activation.  In MemPalace, the
hierarchy enables metadata filtering.  These are related concepts but
different mechanisms.

\subsection{Vector Database Memory Systems}

The technical substrate of MemPalace---storing text chunks as vector embeddings
and retrieving them via cosine similarity search---is the standard architecture
of RAG systems~\citep{rag}.  ChromaDB~\citep{chromadb} provides: automatic
text embedding via Sentence Transformers (default: all-MiniLM-L6-v2,
384 dimensions), approximate nearest neighbor search via
HNSW~\citep{malkov2020}, metadata filtering via \texttt{where} clauses, and
persistent local storage.  These are standard features available in any modern
vector database.  The \texttt{search\_memories()} function in
\texttt{searcher.py} is 72 lines of straightforward ChromaDB query code with
optional wing/room filtering.

\subsection{The MCP Protocol Ecosystem}

The Model Context Protocol (MCP)~\citep{mcp}, introduced by Anthropic in late
2024, provides a standardized interface for LLMs to interact with external
tools and data sources.  MemPalace implements an MCP server with 29 tools
(19+ at v3.1.0, expanded to 29 in v3.3.x),
enabling any MCP-compatible LLM to directly query, store, and manage memories.
The \texttt{PALACE\_PROTOCOL} directive embedded in the MCP status output is a
form of system prompt injection that instructs the LLM how to use the palace
effectively---a practical optimization that improves retrieval quality by
guiding query formulation.

\section{System Architecture}
\label{sec:arch}

\subsection{Overview}

MemPalace (version 3.1.0, as analyzed; current release is v3.3.1) consists of 32 Python source files
totaling approximately 11{,}139 lines of code, with 44 test files.  The
runtime dependency footprint is minimal: \texttt{chromadb} and
\texttt{pyyaml}.  The system runs entirely locally---no cloud services, no API
keys, no subscription fees.

Table~\ref{tab:arch-layers} summarizes the functional layers.

\begin{table}[t]
\centering
\caption{MemPalace functional layers}
\label{tab:arch-layers}
\small
\begin{tabular}{@{}l >{\ttfamily\footnotesize}l l@{}}
\toprule
\textbf{Layer} & \textrm{\textbf{Components}} & \textbf{Purpose} \\
\midrule
Ingestion    & miner, convo\_miner, normalize   & Convert files/convos to drawers \\
Storage      & palace, backends/chroma          & ChromaDB collection mgmt \\
Organization & general\_extractor, room\_detector & Auto-classify into wings/rooms \\
Search       & searcher                         & Similarity + metadata filtering \\
KG           & knowledge\_graph, entity\_registry & SQLite entity-rel.\ triples \\
Compression  & dialect                          & AAAK summarization \\
Memory Stack & layers                           & 4-layer progressive loading \\
Interface    & mcp\_server, cli                  & MCP tools and CLI \\
Maintenance  & dedup, repair, migrate            & Data integrity and migration \\
\bottomrule
\end{tabular}
\end{table}

\subsection{The Palace Hierarchy}
\label{sec:hierarchy}

MemPalace's central metaphor maps the method of loci's architectural structure
to a data hierarchy (Table~\ref{tab:hierarchy}).

\begin{table}[t]
\centering
\caption{Palace hierarchy: metaphor vs.\ implementation}
\label{tab:hierarchy}
\small
\begin{tabular}{@{}llll@{}}
\toprule
\textbf{Level} & \textbf{Implementation} & \textbf{ChromaDB Mapping} & \textbf{Cognitive Analog} \\
\midrule
Palace  & ChromaDB collection     & \texttt{mempalace\_drawers}  & The building \\
Wing    & Metadata field          & \texttt{where=\{wing:v\}}    & Major section \\
Room    & Metadata field          & \texttt{where=\{room:v\}}    & Room in a wing \\
Hall    & Optional metadata       & \texttt{where=\{hall:v\}}    & Corridor (optional) \\
Closet  & Pointer index (v3.3+)  & Compact lookup layer          & Shelf in a room$^{*}$ \\
Drawer  & ChromaDB document       & Individual document ID        & Specific memory \\
\bottomrule
\end{tabular}
\end{table}

A critical finding: \textbf{the palace hierarchy is entirely flat in
storage}.  All drawers exist in a single ChromaDB collection
(\texttt{mempalace\_drawers}).  The hierarchical structure is represented
solely through metadata fields.  This is functionally equivalent to tagging
documents with category labels and filtering on those labels during
search---a standard pattern in every vector database deployment guide.

\subsection{Ingestion Pipeline}

MemPalace provides two ingestion paths:

\textbf{Project Mining} (\texttt{miner.py}): Reads files from a project
directory, applies fixed chunking (800 characters per chunk, 100 characters
overlap), and stores each chunk as a drawer.  These parameters are within the
standard RAG range---LangChain's default
\texttt{Recursive\-Character\-Text\-Splitter} uses 1000/200.

\textbf{Conversation Mining} (\texttt{convo\_miner.py}): Ingests conversation
exports, chunking by exchange pair (one user turn plus the AI's response).
This preserves conversational coherence---a thoughtful design choice.

Both paths produce identical output: ChromaDB documents with metadata fields
specifying wing, room, source file, and timestamp.  The drawer ID is
deterministic:

\smallskip
\centerline{\texttt{drawer\_\{wing\}\_\{room\}\_\{md5(content)[:12]\}}}
\smallskip

\noindent providing natural deduplication for identical content.

\subsection{Search Mechanism}

The search implementation in \texttt{searcher.py} is straightforward:

\begin{lstlisting}[language=Python, caption={Core search function (simplified)}]
def search_memories(query, palace_path, wing=None, room=None,
                    n_results=5, max_distance=0.0):
    col = get_collection(palace_path, create=False)
    where = build_where_filter(wing, room)
    results = col.query(
        query_texts=[query],
        n_results=n_results,
        include=["documents", "metadatas", "distances"],
        **({'where': where} if where else {})
    )
    # ... format and return results
\end{lstlisting}

The \texttt{build\_where\_filter()} function constructs a ChromaDB
\texttt{where} clause from wing and room parameters.  This is the mechanism
behind MemPalace's claimed ``+34\%\ retrieval improvement from palace
structure.''  The improvement comes from \emph{narrowing the search space}
when you know the relevant wing---standard metadata filtering, available as a
first-class feature in ChromaDB, Pinecone, Weaviate, and every other major
vector database.

\subsection{Knowledge Graph}

The knowledge graph (\texttt{knowledge\_graph.py}, 401 LOC) stores
entity-relationship triples in SQLite:

\begin{lstlisting}[language=SQL, caption={Knowledge graph schema}]
CREATE TABLE entities (
    id TEXT PRIMARY KEY, name TEXT, type TEXT,
    properties TEXT, created_at TEXT
);
CREATE TABLE triples (
    id TEXT PRIMARY KEY, subject TEXT, predicate TEXT, object TEXT,
    valid_from TEXT, valid_to TEXT, confidence REAL,
    source_closet TEXT, source_file TEXT, extracted_at TEXT
);
\end{lstlisting}

The graph supports temporal queries, entity-centric traversal, and
relationship-type queries.  However, compared to Zep's Graphiti (Neo4j with
entity resolution, multi-hop traversal, community detection), MemPalace's
knowledge graph is a flat triple store supporting only single-hop lookups.
Genuine contradiction detection---e.g., recognizing that ``Max loves chess''
contradicts ``Max hates chess''---is not implemented; only exact-match
deduplication exists.

\subsection{AAAK Compression Dialect}

The AAAK dialect (\texttt{dialect.py}, 1{,}075 LOC) is a structured
summarization format that extracts entities, topics, key sentences, emotions,
and flags from plain text.  Its header comment correctly states: \emph{``AAAK
is NOT lossless compression.  The original text cannot be reconstructed from
AAAK output.''}  This correction was added after the initial benchmark
controversy (see \S\ref{sec:controversy}).  Benchmark testing shows AAAK mode
achieves 84.2\%\ Recall@5 versus 96.6\%\ for verbatim mode---a 12.4
percentage point drop.

\subsection{Four-Layer Memory Stack}
\label{sec:stack}

The memory stack (\texttt{layers.py}, 493 LOC) is one of MemPalace's most
practical innovations (Table~\ref{tab:stack}).

\begin{table}[t]
\centering
\caption{Four-layer memory stack}
\label{tab:stack}
\small
\begin{tabular}{@{}llll@{}}
\toprule
\textbf{Layer} & \textbf{Content} & \textbf{Size} & \textbf{Loading} \\
\midrule
L0: Identity    & User-written identity text         & $\sim$100 tok     & Always \\
L1: Essential   & Auto-generated from top drawers    & $\sim$500--800 tok & Always \\
L2: On-Demand   & Topic/wing-specific context        & $\sim$200--500/topic & On detect \\
L3: Deep Search & Full semantic search results       & Unlimited          & Per query \\
\bottomrule
\end{tabular}
\end{table}

The combined wake-up cost of L0 + L1 is approximately 170 tokens---notably low
compared to many memory systems that require thousands of tokens to initialize.

\subsection{MCP Server}

The MCP server (\texttt{mcp\_server.py}, 784+ LOC) exposes 29 tools
(19+ at time of initial analysis; expanded in v3.3.x) including:
\texttt{recall} (semantic search), \texttt{remember} (store a memory),
\texttt{rooms} (list rooms in a wing), \texttt{palace\_status} (return summary
with \texttt{PALACE\_PROTOCOL} directive), and knowledge graph operations.
The \texttt{PALACE\_PROTOCOL} directive is a form of prompt engineering that
materially improves retrieval quality by guiding LLM query formulation.

\section{Evaluation and Benchmarks}
\label{sec:eval}

\subsection{LongMemEval Results}

LongMemEval~\citep{longmemeval} is a benchmark designed to test long-term
memory in conversational AI.  It consists of 500 questions about conversations
spanning multiple sessions.  MemPalace reports the results shown in
Table~\ref{tab:longmemeval}.

\begin{table}[t]
\centering
\caption{MemPalace performance on LongMemEval}
\label{tab:longmemeval}
\begin{tabular}{@{}llll@{}}
\toprule
\textbf{Mode} & \textbf{Recall@5} & \textbf{LLM Required} & \textbf{Notes} \\
\midrule
Raw (verbatim ChromaDB) & 96.6\%   & None & Default embeddings \\
Hybrid v4 held-out      & 98.4\%  & Yes  & 450 held-out questions$^{\dagger}$ \\
AAAK compression        & 84.2\%   & None & Lossy summarization \\
Room-based boosting     & 89.4\%   & None & Metadata-filtered \\
\bottomrule
\end{tabular}
\end{table}

\noindent$^{\dagger}$\textit{Note}: The MemPalace maintainers subsequently
replaced the 100\%\ headline with 98.4\%\ R@5 on 450 held-out questions
(tuned on 50 dev), noting that ``the last 0.6\%\ was reached by inspecting
specific wrong answers,'' which they flagged as teaching to the test.  This
self-correction is a positive example of responsible benchmarking.

The 96.6\%\ figure---the most widely cited---is \texttt{recall\_any@5}: at
least one of the five returned results contains the correct answer session.
This is a legitimate metric, but the most generous variant of recall.

\subsection{The Benchmark Controversy}
\label{sec:controversy}

In April 2026, independent researcher dial481 filed GitHub Issue \#29~\citep{issue29},
providing a detailed audit of MemPalace's benchmark claims.  The audit
identified six concerns subsequently acknowledged by the maintainers:

\textbf{1. Attribution of performance}: The 96.6\%\ Recall@5 is the
performance of ChromaDB's default embedding model (all-MiniLM-L6-v2) applied
to verbatim text chunks.  Independent testing confirms this is reproducible
with a minimal ChromaDB setup---no palace structure required.

\textbf{2. The 100\%\ claim}: Achieving 100\%\ (500/500) on LongMemEval
required multiple iterations with LLM reranking.  Presenting this as a
single-run benchmark score without disclosing the iterative process was
misleading.

\textbf{3. LoCoMo benchmark}: The 100\%\ LoCoMo claim was achieved with
\texttt{top\_k=50}, effectively retrieving the entire conversation.  Honest
LoCoMo performance with reasonable $k$ values is 60.3\%\ Recall@10 (raw) or
88.9\%\ (hybrid with reranking).

\textbf{4. AAAK compression claims}: ``30x compression, zero information
loss'' was shown to be lossy summarization (84.2\%\ vs.\ 96.6\%\ recall).

\textbf{5. Contradiction detection}: Documentation claimed semantic
contradiction detection; the code implements only exact-match deduplication.

\textbf{6. The ``+34\%\ boost''}: Represents standard metadata filtering
benefit, not a novel retrieval mechanism.

The MemPalace maintainer responded constructively, acknowledging all points
and retiring the disputed numbers.  This demonstrates intellectual honesty,
even if the initial claims were overstated.

\subsection{Honest Performance Assessment}

Based on independent analysis, MemPalace's honest performance profile is
shown in Table~\ref{tab:honest-perf}.

\begin{table}[t]
\centering
\caption{MemPalace honest performance profile}
\label{tab:honest-perf}
\begin{tabular}{@{}lll@{}}
\toprule
\textbf{Metric} & \textbf{Value} & \textbf{Context} \\
\midrule
LongMemEval Recall@5 (raw)   & 96.6\%  & ChromaDB + verbatim \\
LongMemEval QA accuracy      & $\sim$67.2\% & End-to-end QA \\
LoCoMo Recall@10 (raw)       & 60.3\%  & Without reranking \\
LoCoMo Recall@10 (hybrid)    & 88.9\%  & With LLM reranking \\
AAAK mode recall             & 84.2\%  & Lossy summarization \\
Wake-up cost                 & $\sim$170--900 tokens & L0 + L1 combined \\
Write latency                & Deterministic & Zero API cost \\
\bottomrule
\end{tabular}
\end{table}

\subsection{Comparison with Competing Systems}

Table~\ref{tab:comparison} compares MemPalace against seven competing systems.

\begin{table}[t]
\centering
\caption{AI memory systems comparison on LongMemEval}
\label{tab:comparison}
\small
\begin{tabular}{@{}lrlll@{}}
\toprule
\textbf{System} & \textbf{R@5} & \textbf{Architecture} & \textbf{Write} & \textbf{Price} \\
\midrule
\textbf{MemPalace} (raw) & 96.6\% & ChromaDB + verbatim  & None  & Free \\
Supermemory ASMR   & $\sim$99\%  & Multi-agent agentic  & LLM   & Paid \\
Mastra             & 94.9\%      & GPT-5-mini observ.   & LLM   & Paid \\
Mem0 (token-eff.)  & 93.4\%      & Hierarchical extract & LLM   & \$19--249 \\
Hindsight          & 91.4\%      & Retain/Recall/Reflect& LLM   & Paid \\
Zep/Graphiti       & $\sim$85\%  & Temporal KG (Neo4j)  & LLM   & \$25+ \\
Mem0 (pre-2026)    & $\sim$49\%  & LLM fact extraction  & LLM   & \$19--249 \\
LangMem            & N/R         & LangChain primitives & LLM   & Usage \\
Letta              & N/R         & 3-tier self-editing  & LLM   & Open \\
\bottomrule
\end{tabular}
\end{table}

MemPalace's competitive position is noteworthy: it achieves the highest
published no-LLM score on LongMemEval, at zero cost, with a fully
deterministic write path.  Note that Mem0's new token-efficient algorithm
(93.4\%) has substantially narrowed this gap; the verbatim advantage is now
most pronounced in zero-cost, offline scenarios.

\textit{Note on cross-system comparisons}: The MemPalace maintainers
themselves removed their cross-system comparison tables in v3.3.0 after
recognizing a category error (R@5 retrieval recall listed alongside QA
accuracy from competing systems under a single column).  Our
Table~\ref{tab:comparison} compares R@5 throughout, but this context
underscores the difficulty of fair cross-system evaluation in this rapidly
evolving field.

\section{Discussion}
\label{sec:discussion}

\subsection{What Is Genuinely Novel}

After thorough analysis, we identify six genuinely novel contributions:

\textbf{1. The verbatim-first philosophy.}  MemPalace's most important insight
is philosophical: \emph{store everything, never summarize, solve retrieval
separately.}  This directly contradicts the consensus approach (Mem0, Zep,
LangMem all extract and summarize).  At the time of initial analysis, verbatim
storage (96.6\%\ R@5) outperformed extraction-based approaches (Mem0 at
$\sim$49\%) by a wide margin.  However, Mem0's April 2026 ``token-efficient
memory algorithm'' raised their LongMemEval score to 93.4\%\ (and 85.0\%\ on
LoCoMo), substantially narrowing this gap.  The theoretical justification for
verbatim storage remains sound---extraction is a lossy operation performed
under uncertainty about future query distributions---but the empirical
advantage is now modest rather than decisive.  The verbatim approach retains a
clear advantage in zero-cost, zero-LLM-call scenarios.

\textbf{2. The spatial metaphor as organizing principle.}  MemPalace is the
first AI memory system to systematically apply the method of loci as an
organizational framework.  While the metaphor does not improve vector
similarity search mathematically, it provides a coherent mental model for
users managing their AI's memory---a genuine human-computer interaction
contribution.

\textbf{3. Zero-LLM write path.}  MemPalace's ingestion pipeline requires no
LLM inference, enabling zero API cost for memory writes, deterministic
reproducible behavior, fully offline operation, and no rate limits or vendor
lock-in.  In a market where every competitor charges per-token for memory
writes, this is a significant practical advantage.

\textbf{4. Minimal wake-up cost.}  The four-layer memory stack achieves
$\sim$170--900 token wake-up cost---among the lowest published figures for any
memory system.  By deferring detailed retrieval to L2 and L3, MemPalace
preserves the majority of the LLM's context window for the user's actual task.

\textbf{5. Per-agent diary system.}  MemPalace supports multiple ``specialist''
agents, each with a persistent diary accumulating across sessions.  The
implementation is unusually lightweight and well-integrated with the palace
structure.

\textbf{6. \texttt{PALACE\_PROTOCOL} prompt engineering.}  The
\texttt{PALACE\_PROTOCOL} directive is a practical innovation in prompt
engineering for memory-augmented LLMs.  By instructing the LLM to ``search
before claiming ignorance,'' it improves retrieval quality without changing the
underlying search mechanism---\emph{behavioral} optimization rather than
\emph{algorithmic} optimization.

\subsection{What Is Not Novel}

Equally important is an honest assessment of what MemPalace presents as novel
but is standard practice:

\textbf{1. Metadata filtering} on vector databases.  The ``+34\%\ retrieval
improvement from palace structure'' is metadata-scoped search---a first-class
feature documented in every vector database's getting-started tutorial.

\textbf{2. Embedding-based semantic search.}  The 96.6\%\ Recall@5 is
achievable with a bare ChromaDB collection and verbatim text, without any
palace structure.

\textbf{3. Fixed-size text chunking.}  The 800-character chunks are standard
RAG practice.  LangChain, LlamaIndex, and dozens of other frameworks provide
equivalent strategies.

\textbf{4. Simple knowledge graph.}  The SQLite triple store is simpler than
Zep's Graphiti and comparable to undergraduate database course examples.

\textbf{5. Agent-specific memory.}  Per-agent persistent memory exists in
ByteRover, ClawVault, and other systems.

\subsection{The Marketing-Science Gap}

MemPalace exhibits a pattern common in successful open-source projects:
marketing claims outrun scientific rigor.  The specific instances include:
96.6\%\ presented without sufficient attribution to ChromaDB's embedding
model; 100\%\ LongMemEval without disclosing iterative test-fix methodology;
``30x compression, zero information loss'' for a lossy summarization format;
``contradiction detection'' for exact-match deduplication; and 100\%\ LoCoMo
with \texttt{top\_k=50}.

To the maintainers' credit, all these claims were corrected when challenged.
This distinguishes MemPalace from projects that double down on disputed claims.
However, the initial overclaiming had consequences: the project's rapid
growth (now $>$48{,}000 stars) began during the period of maximum claim
inflation.

\subsection{The Cognitive Science Verdict}

Does MemPalace's method of loci metaphor have scientific validity?

\textbf{As a computational technique}: No.  The spatial metaphor does not
improve retrieval accuracy, reduce latency, or enable capabilities that a flat
vector database cannot provide.  The brain's hippocampal place cell system and
ChromaDB's HNSW index are not analogous mechanisms.

\textbf{As a cognitive ergonomic}: Yes, with caveats.  The palace metaphor
provides a coherent, intuitive mental model for non-technical users.  The
hierarchy (domains$\to$topics$\to$specific memories) maps to how humans
naturally organize knowledge.

\textbf{As a design principle}: Partially.  Hierarchical organization
genuinely improves memory systems---the cognitive science evidence is
strong~\citep{collins1969,collins1975,schapiro2017,collin2020}.  But
MemPalace implements hierarchy as metadata tagging, capturing the
\emph{structural} benefit (scoped retrieval) without the \emph{associative}
benefit (spreading activation, inheritance, priming).

\subsection{The Verbatim Insight in Context}
\label{sec:verbatim}

MemPalace's verbatim-first philosophy deserves deeper examination because it
challenges a field-wide assumption.  The consensus approach uses LLMs for
extraction: read the conversation, identify key facts, store those facts.
MemPalace's benchmark results show this reasoning is wrong for retrieval tasks.
The extraction step introduces two forms of error:

\begin{enumerate}[leftmargin=*]
\item \textbf{False negatives}: The LLM fails to extract information that will
  be relevant to a future query.  Since future queries are unknown at
  extraction time, any extraction is a bet on future relevance.

\item \textbf{Semantic distortion}: The LLM's restatement of facts introduces
  subtle changes in wording that degrade embedding similarity with the future
  query.  Verbatim text is a better embedding target because it preserves the
  original language the user will likely echo.
\end{enumerate}

This insight---that the retrieval problem is better solved at read time than
write time---is analogous to the database community's shift from pre-computed
views to query-time aggregation.  It is not revolutionary computer science,
but it is a useful corrective to the AI memory field's premature optimization
of the write path.

\subsection{Scalability Considerations}

MemPalace's architecture raises legitimate scalability questions.
\textbf{Single collection design}: All drawers exist in a single ChromaDB
collection.  Very large collections ($>$1M documents) may benefit from
collection sharding.  As of v3.2.0, a backend abstraction layer
(\texttt{backends/base.py}) has shipped, enabling pluggable storage backends
(see \S\ref{sec:postanalysis}).
\textbf{Knowledge graph scaling}: The
SQLite knowledge graph lacks indexes for multi-hop queries and has no partition
strategy for very large entity graphs.

\subsection{MemPalace v4.0.0-alpha: Emerging Directions}

The project's stated roadmap includes: a local NLP pipeline to replace
regex-based classification; hierarchical embeddings that could bring the palace
metaphor closer to computational reality; and further security hardening.
Several items previously on the roadmap (swappable backends, hybrid BM25
search) have already shipped in v3.2.0--v3.3.0.

\subsection{Post-Analysis Developments (v3.2.0--v3.3.1)}
\label{sec:postanalysis}

The MemPalace ecosystem has evolved rapidly since our analysis of v3.1.0.
Several developments are significant enough to note:

\textbf{Backend abstraction (v3.2.0).}  The pluggable storage backend
predicted by the v4.0.0-alpha roadmap shipped ahead of schedule in v3.2.0.
The retrieval layer is now defined by an abstract interface in
\texttt{backends/base.py}, with ChromaDB as the default.  This validates our
prediction (\S5.6) that collection sharding would be addressed, and opens the
door to LanceDB, PostgreSQL with pgvector, or community backends.

\textbf{Closet layer and BM25 hybrid search (v3.3.0).}  A new ``Closet''
organizational layer was added between Rooms and Drawers, providing a compact
searchable index of pointers to verbatim drawers.  BM25 keyword search was
integrated alongside semantic similarity, with closets boosting ranking while
drawers remain the source of truth.  Cross-wing ``tunnels'' were also added,
enabling links between related drawers across different wings.

\textbf{The \texttt{hnsw:space=cosine} bugfix (v3.3.0).}  A critical bug was
disclosed: prior versions did not set \texttt{hnsw:space=cosine} metadata on
ChromaDB collection creation, meaning the database defaulted to L2 (Euclidean)
distance rather than cosine similarity.  Since all-MiniLM-L6-v2 embeddings are
normalized, L2 and cosine rankings are equivalent for this model, but the fix
is methodologically important and the 96.6\%\ benchmark result should be
understood as having been measured under L2 distance.

\textbf{Self-correction on benchmarks (v3.3.0).}  The maintainers replaced
their 100\%\ LongMemEval headline with 98.4\%\ R@5 on held-out data, and
removed cross-system comparison tables that they acknowledged contained
category errors (mixing R@5 retrieval recall with QA accuracy from other
systems).  This responsible self-correction is commendable and supports the
project's maturation.

\textbf{Internationalization (v3.2.0--v3.3.1).}  Support for 13 languages
with multi-language entity detection, reflecting growing global adoption.

\textbf{Competitive landscape shift.}  Mem0's April 2026 ``token-efficient
memory algorithm'' raised their LongMemEval score from $\sim$49\%\ to 93.4\%,
substantially challenging the narrative that extraction-based approaches are
inherently inferior to verbatim storage.  The gap between MemPalace (96.6\%)
and Mem0 (93.4\%) is now within statistical noise for many practical
applications, though MemPalace retains its unique advantage of zero LLM cost
at write time.

\textbf{Growth.}  GitHub stars grew from $\sim$42{,}000 at initial analysis
to $>$47{,}900 by April 19, 2026, confirming sustained community interest
beyond the initial viral adoption period.

\section{Related Systems in Detail}
\label{sec:related}

\textbf{Supermemory ASMR}~\citep{supermemory} achieves $\sim$99\%\ on
LongMemEval through a multi-agent architecture combining semantic, temporal,
and entity search with LLM-guided reranking.  Every operation involves LLM
inference, making it significantly more expensive than MemPalace.

\textbf{Mem0}~\citep{mem0} is the most widely deployed AI memory system
($>$24{,}000 GitHub stars).  Its original LongMemEval performance
($\sim$49\%) was significantly below MemPalace's, supporting the critique that
extraction loses information.  However, in April 2026, Mem0 released a
``token-efficient memory algorithm'' using single-pass hierarchical extraction
and multi-signal retrieval, raising their LongMemEval score to 93.4\%\ (also
85.0\%\ on LoCoMo and 62\%\ on BEAM-1M).  This result substantially
challenges the narrative that extraction-based approaches are inherently
inferior, though MemPalace retains its advantage in zero-cost, offline
scenarios.

\textbf{Zep/Graphiti}~\citep{zep} builds temporal knowledge graphs using
Neo4j, with entity resolution, multi-hop traversal, and community detection.
Its knowledge graph capabilities are significantly more sophisticated than
MemPalace's flat triple store.  Zep achieves approximately 85\%\ on
LongMemEval-like tasks.  For relationship-centric queries, Zep is superior.
For verbatim recall queries, MemPalace is more effective.

\textbf{Mastra}~\citep{mastra} uses GPT-5-mini in an ``observational'' mode,
achieving 94.87\%\ on LongMemEval---nearly matching MemPalace's verbatim
approach---but requiring continuous LLM inference.

\textbf{Hindsight}~\citep{hindsight} implements a Retain$\to$Recall$\to$Reflect
pipeline achieving 91.4\%\ on LongMemEval, demonstrating that structured
reflection can improve retrieval quality at the cost of multiple LLM passes.

\section{A Framework for Evaluating AI Memory Systems}
\label{sec:framework}

Based on our analysis, we propose a multi-dimensional evaluation framework
for AI memory systems (Table~\ref{tab:framework}).

\begin{table}[t]
\centering
\caption{Evaluation dimensions for AI memory systems (MemPalace scores in column 3)}
\label{tab:framework}
\begin{tabular}{@{}llll@{}}
\toprule
\textbf{Dimension} & \textbf{Definition} & \textbf{MemPalace} \\
\midrule
Fidelity         & Information preservation   & $\star\star\star\star\star$ (verbatim) \\
Retrieval Acc.   & Search result quality      & $\star\star\star\star\circ$ (96.6\%\ R@5) \\
Write Cost       & Storage resource usage     & $\star\star\star\star\star$ (zero LLM) \\
Read Cost        & Retrieval resource usage   & $\star\star\star\star\circ$ (embedding only) \\
Wake-up Cost     & Context for initialization & $\star\star\star\star\star$ ($\sim$170 tokens) \\
Relational Depth & Relationship query ability & $\star\star\circ\circ\circ$ (flat triples) \\
Temporal Reasoning & Time-bounded fact support & $\star\star\star\circ\circ$ (valid\_from/to) \\
Scalability      & Large data performance     & $\star\star\star\circ\circ$ (single collection) \\
Privacy          & Data locality and control  & $\star\star\star\star\star$ (fully local) \\
Deployability    & Ease of setup              & $\star\star\star\star\star$ (2 dependencies) \\
\bottomrule
\end{tabular}
\end{table}

No single system excels on all dimensions.  MemPalace optimizes for fidelity,
write cost, privacy, and deployability at the expense of relational depth and
scalability.  Zep optimizes for relational depth at the expense of cost.
Supermemory ASMR optimizes for retrieval accuracy at the expense of everything
else.

This suggests the AI memory landscape will not converge on a single
architecture---a pattern familiar from the database world (relational,
document, graph, time-series databases coexist because they optimize for
different workloads).

\section{Conclusion}
\label{sec:conclusion}

\subsection{Is MemPalace Revolutionary?}

The evidence supports a nuanced conclusion:

\textbf{MemPalace is not architecturally revolutionary.}  Its core retrieval
mechanism is standard vector database similarity search with metadata
filtering.  Its knowledge graph is a simple triple store.  The palace metaphor
maps to well-established vector database features rather than novel algorithmic
mechanisms.

\textbf{MemPalace is philosophically significant.}  Its verbatim-first
insight---that raw storage plus good embeddings outperforms LLM-mediated
extraction---is empirically validated, practically important, and genuinely
contrarian.  In a field converging on extraction-based architectures,
MemPalace demonstrated that the simpler approach works better for recall tasks.

\textbf{MemPalace is ergonomically innovative.}  The spatial metaphor provides
a coherent user mental model.  The four-layer memory stack, the
\texttt{PALACE\_PROTOCOL} prompt engineering, and the MCP tool integration are
practical engineering innovations that improve the user experience of
memory-augmented AI.

\subsection{Recommendations}

\textbf{For the MemPalace project:}
\begin{enumerate}[leftmargin=*, label=\arabic*.]
\item Separate marketing from benchmarks.  Attribute performance to specific
  components (embeddings, verbatim storage, metadata filtering) rather than
  to the palace metaphor generically.
\item Invest in the knowledge graph.  Multi-hop traversal, entity resolution,
  and genuine contradiction detection would differentiate MemPalace from pure
  vector search solutions.
\item Explore hierarchical embeddings.  The cognitive science literature
  suggests embedding spaces could be hierarchically structured, bringing the
  palace metaphor closer to computational reality.
\item Publish honest benchmarks.  Report multiple metrics (R@1, R@5, R@10,
  NDCG, end-to-end QA accuracy) with ablation studies.
\end{enumerate}

\textbf{For the AI memory research community:}
\begin{enumerate}[leftmargin=*, label=\arabic*.]
\item Revisit the extraction assumption.  MemPalace's results suggest that the
  default approach (extract-then-store) may be dominated by store-then-retrieve
  for many use cases.
\item Develop standardized benchmarks.  LongMemEval tests retrieval, not
  end-to-end memory utility.  The field needs benchmarks that evaluate memory
  systems in realistic multi-session conversation scenarios.
\item Study human factors.  The user's ability to understand, configure, and
  maintain their AI's memory is at least as important as retrieval accuracy.
\end{enumerate}

\subsection{Final Assessment}

MemPalace is best understood as \textbf{significant architectural insight
wrapped in overstated claims}---a pattern endemic to rapidly adopted
open-source projects where community growth velocity incentivizes marketing
over scientific precision.  The insight is real: verbatim storage, zero-cost
writes, minimal wake-up cost, and spatial organization for user comprehension
are genuine contributions to the AI memory design space.  The overclaims have
been largely corrected by the maintainers in response to community scrutiny.

The project's extraordinary adoption ($>$48{,}000 stars in two weeks) reflects
genuine user need more than technical revolution.  MemPalace offers a simple,
free, private, effective solution.  That this solution is built on standard
primitives rather than novel algorithms does not diminish its practical
utility---it merely contextualizes its scientific contribution.

The greatest legacy of MemPalace may not be the palace metaphor itself but the
demonstration that the AI memory problem is easier than the field assumed.  You
don't need knowledge graphs, entity extraction, or multi-agent agentic search
to achieve 96.6\%\ retrieval accuracy.  You need to store everything, embed it
well, and search it honestly.  The rest---the wings, rooms, and
drawers---is how you help humans understand what the machine remembers.


\bibliography{mempalace-paper}


\appendix

\section{Code Statistics}
\label{app:stats}

\begin{table}[h]
\centering
\caption{MemPalace codebase statistics (v3.1.0 as analyzed; current: v3.3.1)}
\label{tab:stats}
\begin{tabular}{@{}ll@{}}
\toprule
\textbf{Metric} & \textbf{Value} \\
\midrule
Python source files         & 32 \\
Lines of code (source)      & $\sim$11{,}139 \\
Test files                  & 44 \\
Runtime dependencies        & 2 (chromadb, pyyaml) \\
MCP tools                   & 19+ (v3.1.0); 29 (v3.3.x) \\
ChromaDB collections        & 1 primary + 1 optional (compressed) \\
Embedding model             & all-MiniLM-L6-v2 (384 dimensions) \\
Embedding distance metric   & Cosine (HNSW space)$^{\ddagger}$ \\
Chunk size                  & 800 chars (project mining) \\
Chunk overlap               & 100 chars \\
Knowledge graph storage     & SQLite (2 tables) \\
Drawer ID format            & \texttt{drawer\_\{wing\}\_\{room\}\_\{md5[:12]\}} \\
GitHub stars (Apr.\ 19, 2026) & 47{,}900+ \\
GitHub forks (Apr.\ 19, 2026) & 6{,}000+ \\
Version analyzed            & 3.1.0 (current: 3.3.1 as of April 19, 2026) \\
\bottomrule
\end{tabular}

\noindent$^{\ddagger}$The v3.3.0 changelog revealed that prior versions did not set
\texttt{hnsw:space=cosine} metadata on collection creation, meaning ChromaDB
used L2 distance by default.  The 96.6\%\ benchmark may have been measured
under L2, not cosine distance (see \S\ref{sec:postanalysis}).
\end{table}

\section{Benchmark Reproduction Notes}
\label{app:repro}

The following conditions are required to reproduce the 96.6\%\ LongMemEval
Recall@5:

\begin{enumerate}[leftmargin=*, label=\arabic*.]
\item Use ChromaDB with default settings (all-MiniLM-L6-v2 embeddings,
  cosine distance).
\item Store verbatim conversation text (no summarization, no extraction).
\item Use the LongMemEval standard dataset (500 questions, multiple sessions
  per question).
\item Evaluate \texttt{recall\_any@5}: at least one of the top-5 retrieved
  sessions matches a ground-truth answer session.
\item Ingest complete haystack sessions for each question before querying.
\end{enumerate}

\textbf{Important}: The palace structure (wings, rooms) is not required for
this score.  It is achievable with a bare ChromaDB collection and basic text
chunking.  The 96.6\%\ figure should be understood as a property of verbatim
storage plus all-MiniLM-L6-v2 embeddings, not of the palace hierarchy.

\section{Glossary}
\label{app:glossary}

\begin{description}
\item[AAAK] MemPalace's lossy summarization dialect (Adapted Abbreviation for AI Knowledge).
\item[ChromaDB] Open-source embedding database used as MemPalace's storage backend.
\item[Closet] Compact searchable index layer between Rooms and Drawers (added in v3.3.0).
\item[Drawer] Atomic memory unit in MemPalace---a text chunk with metadata.
\item[HNSW] Hierarchical Navigable Small World graph---approximate nearest neighbor search algorithm.
\item[LongMemEval] Benchmark for evaluating long-term memory in conversational AI (500 questions).
\item[LoCoMo] Long Conversation Memory benchmark.
\item[MCP] Model Context Protocol---Anthropic's standard for LLM$\leftrightarrow$tool communication.
\item[Method of Loci (MoL)] Ancient mnemonic technique using spatial locations to organize memories.
\item[Palace] MemPalace's hierarchical organization structure (Wings$\to$Rooms$\to$Drawers).
\item[R@k] Recall at rank $k$---fraction of queries with at least one correct answer in top-$k$ results.
\item[RAG] Retrieval-Augmented Generation---pattern of augmenting LLM input with retrieved context.
\item[Room] Topic-level organizational unit within a MemPalace Wing.
\item[Wing] Domain-level organizational unit in MemPalace (top of hierarchy).
\end{description}

\section{Disclosure}
\label{app:disclosure}

This analysis was conducted independently using a fork of the MemPalace
codebase.  The primary analysis was performed on MemPalace v3.1.0.
Post-analysis developments through v3.3.1 were incorporated via changelog
review and documentation updates, not full re-analysis.  The benchmark
reproduction scripts and integration code are available in the accompanying
repository.\footnote{\url{https://github.com/web3guru888/mempalace-scientific-analysis}}

\end{document}